# Masked Face Recognition under Different Backbones

Bo Zhang, Ming Zhang
TravelSky Technology
Limited,101300
Beijing, China
zhangb@travelsky.com.cn
mingzh@travelsky.com.cn

Kun Wu, Lei Bian
TravelSky Mobile Technology
Limited, 100087
Beijing, China
wk@travelsky.com.cn
bianlei@travelsky.com.cn

Yi Lin
College of Computer Science
Sichuan University, 610000
Chengdu, China
yilin@scu.edu.cn

*Purpose.* Erratum to the paper (Zhang et al., 2025): corrections to Table IV and the data in Page 3, Section A.

*Abstract*—In the post-pandemic era, a high proportion of civil aviation passengers wear masks during security checks, posing significant challenges to traditional face recognition models. The backbone network serves as the core component of face recognition models. In standard tests, r100 series models excelled (98%+ accuracy at 0.01% FAR in face comparison, high top1/top5 in search). r50 ranked second, r34_mask_v1 lagged. In masked tests, r100_mask_v2 led (90.07% accuracy), r50_mask_v3 performed best among r50 but trailed r100. Vit-Small/Tiny showed strong masked performance with gains in effectiveness. Through extensive comparative experiments, this paper conducts a comprehensive evaluation of several core backbone networks, aiming to reveal the impacts of different models on face recognition with and without masks, and provide specific deployment recommendations.

## I. Introduction

Facial recognition technology is increasingly employed as a biometric identification method in various access-control and security contexts—including airports, government buildings, and high-security facilities, as well as for surveillance tasks such as criminal investigations [1–4].

In recent years, rapid growth in global air travel and steadily rising passenger throughput have placed unprecedented demands on airport operational efficiency, safety, and service quality. At the same time, travelers' habits in information access, mobility, and spending have undergone profound changes, making traditional airport service models insufficient to meet diverse needs and creating an urgent imperative to embrace intelligent, innovative solutions. High-tech verification methods such as facial recognition are being adopted in airports worldwide, not only to greatly accelerate security checkpoints but also—thanks to their "precise identity authentication"—to serve as aviation's first line of defense, effectively preventing identity fraud and unauthorized access, aiding law-enforcement agencies in tracking criminals and identifying wanted persons, and enabling the rapid recovery of lost individuals and unaccompanied minors. IATA's One ID initiative [5] aims to leverage two pillars, "digitalization of admissibility" and "contactless travel", so that travelers can complete visa, health-status, and other eligibility checks online before arriving at the airport and then pass through check-in, baggage drop, security, and boarding with a single live-face verification, entirely removing the need to present documents multiple times, truly achieving "Ready to Fly."

As mask wearing has become the norm, traditional face-recognition systems face increased information loss, placing higher demands on real-time performance and robustness; therefore, developing fast, highly accurate masked-face recognition technology has become crucial for ensuring public safety and enhancing the passenger experience. In particular, systematically evaluating different backbone architectures, such as convolutional networks and vision transformers, allows us to understand trade-offs between feature representation under occlusion, computational efficiency, and inference speed. By selecting and tuning the optimal backbone for masked-face recognition, we can significantly boost the system's ability to extract discriminative features from partially covered faces, ensuring both high accuracy and the low latency required in busy airport environments.

The remainder of this paper is organized as follows. Section II surveys the existing literature, critically appraising prior approaches and identifying the gaps that this research aims to address. Section III describes the proposed model and the methodological framework. Section IV presents the validation of the model through a series of experiments and analyses, demonstrating its effectiveness and robustness under various conditions. Finally, Section V summarizes the principal findings, discusses their implications, and outlines directions for future research.

## II. Literature Review

Recent years have seen significant progress in both mask detection and masked face recognition. For mask detection, Loey et al. (2021) introduced a YOLO-v2 detector with a ResNet-50 backbone, achieving an average precision of 81% on a combined Medical Masks and Face Mask dataset by leveraging deep feature extraction and one-stage detection architecture [6]. Building on this, Kumar et al. (2021) enhanced tiny YOLO v4 with a Spatial Pyramid Pooling (SPP) module to better handle small mask regions, reporting an improved precision of 84% on their self-created mask dataset [7].

With robust mask detection in place, multiple strategies for recognizing masked faces have emerged. Mandal et al. (2021) fine-tuned a pre-trained ResNet-50 on unmasked faces and then adapted it to masked inputs, exploring occlusion cropping and supervised domain adaptation to mitigate feature loss [8]. Li et al. (2021) proposed a cropping-and-attention scheme that focuses on the eye region, integrating Convolutional Block Attention Modules into each ResNet-50 block to refine feature maps around critical facial landmarks [9]. Boutros et al. (2021) developed an Embedding Unmasking Model using a Self-Restrained Triplet loss to enforce tighter embedding clusters for masked and unmasked pairs [10], while Deng et al. (2021) presented "MFCosface," which restores masked regions before applying a large margin cosine loss for recognition [11]. More recent work includes Li and Ge's (2020) end-to-end de-occlusion distillation framework that imparts amoral completion capabilities to existing models [12], and Din et al. (2020), who leveraged a dual-discriminator GAN to learn both global facial structure and local occluded details for unmasking [13]. Shahzad et al. (2024) highlight the limitations of unimodal facial expression recognition (FER) under mask occlusion and propose a multimodal deep learning framework that fuses visual cues from a masked-face dataset (M-LFW-F) with vocal features from CREMA-D [14]. Huang et al. (2024) provide a **comprehensive survey** of deep-learning–based masked face recognition (MFR), organizing recent advances according to network architectures (e.g., CNNs, attention models) and feature-extraction strategies (e.g., margin losses, feature alignment)[15]. Liao, Guha, and Sanchez (2025) employs a Random Mask Attention GAN (RMAGAN) that learns to "fill in" missing regions caused by large pose angles without paired frontal images or external renderers[16].

Early backbone networks were centered on convolutional neural networks (CNNs), evolving from LeNet-5's simple convolutional layers into much deeper architectures[17]. Subsequently, AlexNet introduced ReLU activation, data augmentation, and dropout, significantly improving the model's generalization ability[18]. VGGNet stacked successive small (3 × 3) convolutional kernels, enhancing feature representation by increasing network depth[19]. In contrast to VGG, GoogleNet further expanded both depth and width while keeping parameter counts much lower, paving the way for one-stage detectors like YOLO v1[20-21]. With the success of Transformers in natural language processing, Vision Transformer (ViT) was the first to bring self-attention into the vision domain. ViT divides an image into a sequence of patches and models long-range dependencies via global attention, thereby overcoming CNNs' limited local receptive fields[22]. Pyramid Vision Transformer (PVT) builds on this by introducing a hierarchical feature-pyramid structure to optimize multi-scale feature fusion[23]. Transformer-based backbones excel at capturing global context without inductive biases, showing performance that surpasses traditional CNNs in tasks such as image classification and object detection. Hybrid architectures, such as CoAtNet and BoTNet—further improve backbone practicality and efficiency by combining convolutional and attention mechanisms[24-25].

III. MODELS AND BACKBONES

We mainly introduces the models and their backbones in this section, such as the Vision Transformer (ViT) model, the ViT-tiny model, and the ViT-small model.

As discussed in the second section, visual algorithms constructed with different backbones can lead to two distinct outcomes. Firstly, larger models generally offer better recognition efficiency, but they also demand higher specifications from edge devices during deployment. Secondly, some newer and highly efficient models can achieve satisfactory prediction results with relatively fewer parameters.

To thoroughly test different backbones in the fourth chapter, we introduce backbone models with various Vision Transformer architectures and compare them with traditional models.

**Standard ViT Model**: The standard Vision Transformer (ViT) model takes tokens as inputs, similar to how the Transformer processes one-dimensional data. It maps the vectorized patches through a trainable linear projection to the model dimension d, as in:

$$\zeta = [\xi; \xi^1 E; \xi^2 E;...; \xi^N E] + E_\pi \qquad (1)$$

The main architecture of this model is shown in Figure (1).

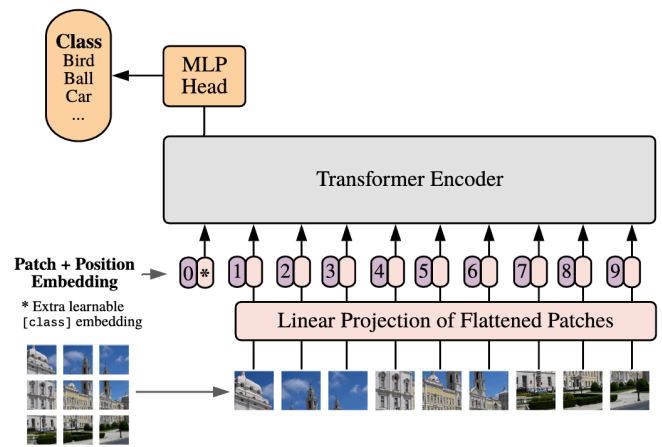

Fig. 1. The Vision Transformer (ViT) Model Overview.

The ViT model can handle sequences of any length, provided that there is sufficient memory available.

**ViT-small Model**: However, the ViT model still has some minor issues, such as the scattering of attention and the ineffective embedding of numerical tokens. The Microsoft research team proposed the Vision Transformer for Small-size, abbreviated as ViT-small. To address the issue of scattered attention, ViT-small introduced Local Self-Attention. In ViT-small, by learning the temperature parameter of the SoftMax function, attention is applied locally. Regarding the problem of ineffective embedding of numerical tokens, ViT-small proposed the Shifted Patch Tokenization, abbreviated as the SPT module. In ViT-small, the image is shifted by half a patch size in each of the four directions of the network and then concatenated with the original image. This enables the model to be trained from scratch even on small-scale datasets.

**ViT-tiny Model**: To further compress the model and enhance the utilization of data from small datasets, researchers proposed the ViT-tiny model. In fact, they employed a method of fast knowledge distillation. With the guidance of a large model, the small model can also benefit from the advantages that the large model gains on large-scale datasets. In principle, ViT-tiny pre-generates some fine-grained probability vectors as "soft" labels for image inputs.

Combined with data augmentation methods such as Random Cropping, RandAugment, and CutMiX, it directly uses these "soft" labels for training. The advantage of this approach is that it allows ViT-tiny to achieve several benefits, including a faster convergence speed and the flexibility for model expansion. Specifically, during each iteration of the training process of ViT-tiny, the model only needs to retrieve the "soft" labels from the stored files and simultaneously optimize the following objective function, as in:

$$\Lambda = XE(\psi, \Omega(A(\xi))) \qquad (2)$$

This step is a typical knowledge distillation process. During the knowledge distillation process, ViT-tiny only smooths the fine-grained labels. For the part of sparse labels, the smoothing method of the model is the inverse function of the sum of the first K ordinary labels.

It is expected that the above three ViT models outperform the ResNet model. However, there are relatively few test results in the complex scenario of face recognition with masks. In this article, in the fourth chapter, we will conduct horizontal comparative tests for both normal face recognition and face recognition with masks. In the fifth chapter, we will elaborate on the performance of each different model and present our test conclusions.

## IV. NUMERICAL RESULTS FOR DIFFERENT BACKBONES

The training dataset employed in this study is webface42M, with the proportion of masked faces set at 15%. Contained 100,000 live ID samples were incorporated into the training process. This dataset combination aims to enhance the model's generalization ability across different facial scenarios, enabling it to recognize both masked and unmasked faces effectively.

### A. ResNet backbone

R100-V1 networks training results are show in Figure (2).

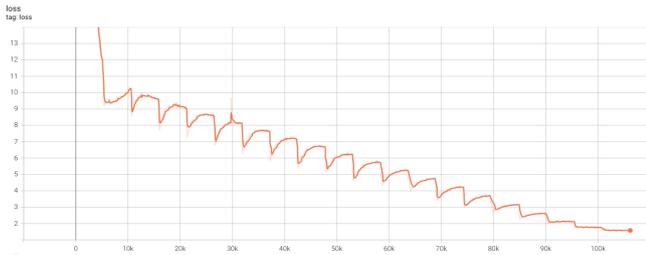

Fig. 2.  Loss function in r100-v1.

r100-V1: The configuration parameters were set as follows: config.margin_list = (1.0, 0.0, 0.4), config.network = "r100". The learning rate started at 0.3 and gradually decayed. The loss function initially decreased linearly with the learning rate but ultimately stabilized at 1.558. In the testing phase, for 10,000 pairs of standard face data (without masks), the threshold at a 0.01% False Acceptance Rate (FAR) was 0.2957, and the accuracy was 98.82%, showing a 0.09% decrease compared to the previous r100 face model. For 100,000 pairs[1] of masked face data, the threshold at a 0.01% FAR was 0.2993, and the accuracy was 88.91%, representing an 8% increase. Analyzing the training process, the relatively high final loss value indicated that the model might not have been fully trained. Thus, considerations for further training included increasing the number of epochs and reducing the learning rate. r100-V2 networks training results are show in Figure (3).

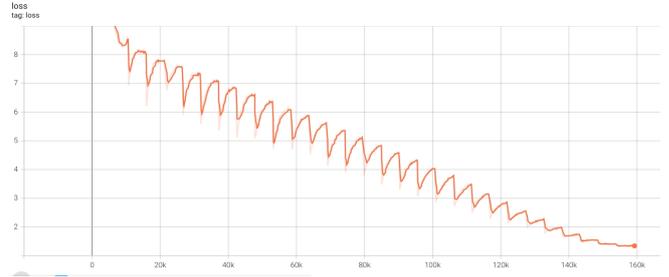

Fig. 3.  Loss function in r100-v2.

Compared to V1, the main change was reducing the initial learning rate to 0.2, and increasing the number of epochs to 30. The loss function decreased to 1.338. In testing, for standard face data, the threshold at a 0.01% FAR was 0.2936, with an accuracy of 98.86%, a 0.04% improvement over r100_mask_v1. For masked face data, the threshold was 0.2988, and the accuracy was 89.95%, a 1% increase. The loss function curve showed significant oscillations in the early epochs due to the relatively high learning rate, suggesting that further reducing the learning rate below 0.1 might be beneficial for better training.

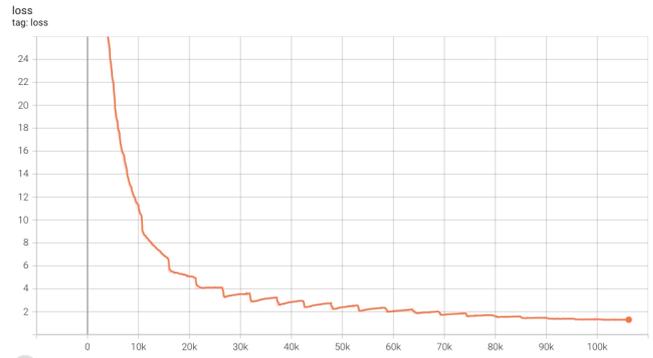

Fig. 4.  Loss function in r100-v3.

r100 V3 Version: As shown in Figure (4). The learning rate was further decreased to 0.03, and the number of epochs was set to 20. The loss function curve was smooth, with a final value of 1.29. For standard face data, the threshold at a 0.01% FAR was 0.3012, and the accuracy was 98.97%, a 0.11% improvement over r100_mask_v2. However, for masked face data, the threshold was 0.3052, and the accuracy was 89.32%, a 0.63% decrease compared to r100_mask_v1. The analysis suggested that the model might have overfitted, and increasing the proportion of masked images to over 20% could potentially improve the recognition of masked faces.

---

[1] The reported value is incorrect due to a data error. The value "10000" should be corrected to "100000."

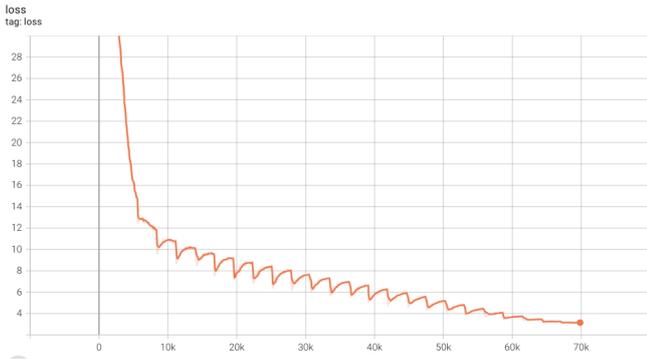

Fig. 5. Loss function in r34-v1.

With parameters like config.network = "r34" and config.lr = 0.3, the model was trained for 25 epochs. As shown in Figure (5), although the loss function curve showed a trend of convergence, specific test results for masked and unmasked face recognition were not fully presented in the original.

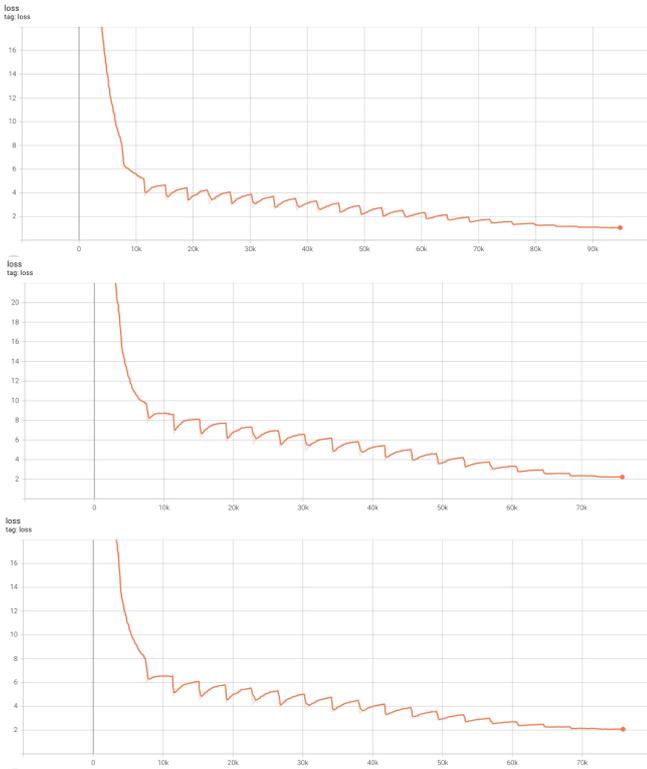

Fig. 6. Loss function in r50-v1(up), v2(middle), and v3(bottom).

As shown in Figure (6), we have: r50 v1 Version: Trained with config.lr = 0.1 for 30 epochs without masked data in the training set. The loss function converged to 1.05 as expected. For standard face data, the threshold at a 0.01% FAR was 0.2978, and the accuracy was 98.303%, a 1.4% improvement over the previous r50 encoding model. For masked face data, the threshold was 0.3014, and the accuracy was 72.569%, a 7.3% increase. However, the relatively low accuracy for masked faces was attributed to the lack of masked data in training. r50 v2 Version: By adding masked data to the training set and adjusting the learning rate to 0.2, the loss function converged to 2.15. In testing, for standard face data, the threshold at a 0.01% FAR was 0.2955, and the accuracy was 98.151%, a 0.15% decrease compared to r50_v1. For masked face data, the threshold was 0.2982, and the accuracy was 84.8%, a significant 12.3% increase. r50 v3 Version: With a learning rate of 0.1 and 20 epochs of training, the loss function converged below 2. The test results showed a slight improvement in the recognition accuracy of standard face data and a marginal increase in the recognition of masked face data compared to r50_mask_v2.

### B. Vit-Tiny Model

Compared with ResNet model, we have the result in Figure (7).

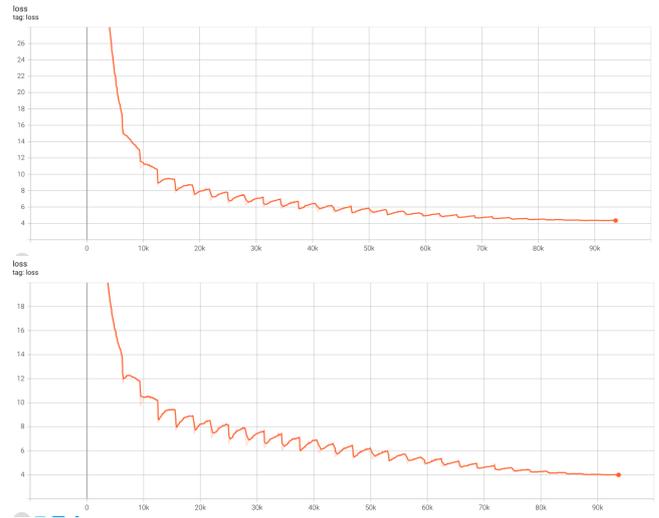

Fig. 7. Loss function in Vit-Tiny-v1(up), and v2(bottom).

Vit-Tiny-v1: Using a relatively low learning rate of 0.001, the loss function curve was smooth but converged at 4.32, indicating that the learning rate might have been set too low. In testing, for 100,000 pairs of standard face data, the threshold at a 0.01% FAR was 0.3558, and the accuracy was 96.228%. Test results for masked face data were incomplete. Vit-Tiny-v2: Increasing the learning rate to 0.003, the loss function converged to 3.96. For standard face data, the threshold at a 0.01% FAR was 0.3556, and the accuracy was 96.887%. Similar to v1, the test results for masked face data were not fully presented.

### C. ResNet backboneVit-Small Model

Configured with config.lr = 0.002 and trained for 30 epochs, the loss function decreased during training.

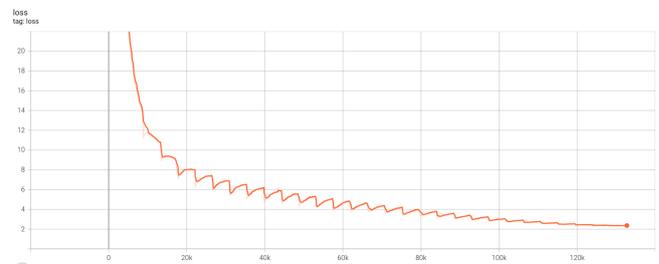

Fig. 8. Loss function in Vit-Small model.

The result is shown in Figure (8). Test results for both standard and masked face data were presented, but a detailed analysis was lacking.

## V. Face Recognition with and without Masks: A Comprehensive Analysis

In this section, we conduct separate tests on the performance of face recognition models for masked and unmasked faces under different backbones. In this test, the training and test data is from Standard Webface42M, with 15% data covered with mask.

### A. The numerical tests for unmasked face recognition

In the unmasked face tests, we carried out comprehensive evaluations for both face comparison and face search.

For the first part, we conducted the face comparison test. The test results are presented in Table I.

TABLE I.  FACE COMPARISON TEST RESULTS FOR 100,000 PAIRS OF UNMASKED FACE TEST DATA

| Backbone | Model Performance | |
|---|---|---|
| | FAR@0.01% Threshold | Accurate Rate (%) |
| R100 | 0.2996 | 99.11 |
| R100-mask-v2 | 0.2991 | 99.06 |
| R100-mask-v3 | 0.2983 | 98.97 |
| R50 | 0.3255 | 96.93 |
| R50-v1 | 0.2978 | 98.30 |
| R50-mask-v2 | 0.2955 | 98.15 |
| R50-mask-v3 | 0.2960 | 98.22 |
| R34-mask | 0.3041 | 97.02 |
| Vit-Tiny-mask | 0.2904 | 99.02 |
| Vit-Small-mask | 0.2994 | 98.86 |

Table I presents the performance of models under different backbones in scenarios without masks. The performance is evaluated using two metrics: FAR@0.01% Threshold (the threshold corresponding to a False Acceptance Rate of 0.01%) and Accurate Rate (in percentage terms).

The values of these two metrics differ across various backbone networks. The R100 backbone network attains the highest accuracy rate, reaching 99.11%. On the other hand, the R50 backbone network has a relatively lower accuracy rate, which is 96.93%. The values of the FAR@0.01% Threshold metric also show fluctuations, depending on the specific backbone network.

TABLE II.  FACE SEARCH RESULT WITH UNMASK UNDER DIFFERENT SEARCH SIZE (TOP1 PERFORMANCE)

| Backbone | Accurate Rate (%) under Different Search Size (*10000) | | | |
|---|---|---|---|---|
| | 1 | 3 | 5 | 10 |
| R100 | 98.18 | 97.40 | 96.96 | 96.15 |
| R100-mask-v2 | 98.15 | 97.32 | 96.85 | 96.05 |
| R100-mask-v3 | 98.06 | 97.11 | 96.60 | 95.73 |
| R50 | 96.08 | 94.17 | 93.11 | 91.39 |
| R50-v1 | 96.99 | 94.96 | 93.28 | 91.06 |
| R50-mask-v2 | 96.54 | 95.01 | 92.65 | 93.19 |
| R50-mask-v3 | 97.26 | 95.13 | 95.13 | 93.84 |
| R34-mask | 96.27 | 94.32 | 93.31 | 91.65 |
| Vit-Tiny | 98.44 | 97.96 | 97.33 | 97.02 |
| Vit-Small | 98.53 | 97.88 | 97.40 | 96.98 |

TABLE III.  FACE SEARCH RESULT WITH UNMASK UNDER DIFFERENT SEARCH SIZE (TOP5 PERFORMANCE)

| Backbone | Accurate Rate (%) under Different Search Size (*10000) | | | |
|---|---|---|---|---|
| | 1 | 3 | 5 | 10 |
| R100 | 99.81 | 99.59 | 99.44 | 99.18 |
| R100-mask-v2 | 99.79 | 99.60 | 99.43 | 99.15 |
| R100-mask-v3 | 99.78 | 99.56 | 99.39 | 99.07 |
| R50 | 99.26 | 98.46 | 97.96 | 97.10 |
| R50-v1 | 99.13 | 98.65 | 98.60 | 97.77 |
| R50-mask-v2 | 99.65 | 99.03 | 98.87 | 98.80 |
| R50-mask-v3 | 99.62 | 99.15 | 98.83 | 98.25 |
| R34-mask | 99.33 | 98.60 | 98.12 | 97.23 |
| Vit-Tiny | 99.96 | 99.64 | 99.55 | 99.51 |
| Vit-Small | 99.92 | 99.73 | 99.63 | 99.55 |

In the second part, we conducted the face search test. The top-1 metric results are shown in Table II, and the top-5 metric results are presented in Table III, respectively.

For the Top1 face search scenario, it can be observed that as the search size increases, the accuracy rates corresponding to various backbone networks generally show a downward trend. For instance, when the search size is 1, the accuracy rate of the R100 backbone network is 98.18%. When the search size increases to 10, the accuracy rate drops to 96.15%. Among these backbone networks, Vit-Tiny and Vit-Small demonstrate relatively high accuracy rates across all search sizes, showcasing excellent performance. In contrast, the R50 backbone network has relatively lower accuracy rates under different search sizes, resulting in subpar performance.

Similarly, in the Top5 face search scenario, as the search size expands, the accuracy rates exhibit a decreasing trend. Take the R100 backbone network as an example. Its accuracy rate is 99.81% when the search size is 1 and decreases to 99.18% when the search size reaches 10. Among these backbone networks, Vit-Tiny stands out with notably high accuracy rates across all search sizes, having a distinct advantage. The R50 backbone network has relatively lower accuracy rates for some search sizes, indicating slightly weaker performance compared to other backbone networks.

### B. The numerical tests for masked face recognition

Furthermore, we tested the performance of models for masked faces. We also conducted comprehensive tests for both face comparison and face search.

TABLE IV. FACE COMPARISON TEST RESULTS FOR 100,000 PAIRS OF MASKED FACE TEST DATA[2]

| Backbone | Performance | |
|---|---|---|
| | FAR@0.01% Threshold | Accurate Rate (%) |
| R100 | 0.3064 | 80.93 |
| R100-mask-v1 | 0.2963 | 88.98 |
| R100-mask-v2 | 0.2986 | 90.07 |
| R100-mask-v3 | 0.3009 | 89.70 |
| R50 | 0.3230 | 65.12 |
| R50-v1 | 0.3014 | 72.57 |
| R50-mask-v2 | 0.2982 | 84.80 |
| R50-mask-v3 | 0.2985 | 85.00 |
| R34-mask | 0.3043 | 79.87 |
| Vit-Tiny-mask | 0.2999 | 89.43 |
| Vit-Small-mask | 0.3041 | 87.64 |

For the first part, we conducted the face comparison test. The test results are presented in Table IV.

We found that due to the obstruction of masks, a large amount of facial information was lost, leading to a significant decline in the performance of various models. Meanwhile, we also noted that models specifically designed for masked-face scenarios performed significantly better than general purpose models. On the other hand, Vit-Tiny and Vit-Small still maintained high levels of model accuracy.

In the second part, we conducted the face search test. The top-1 metric results are shown in Table V, and the top-5 metric results are presented in Table VI, respectively.

TABLE V. FACE SEARCH RESULT WITH MASK UNDER DIFFERENT SEARCH SIZE (TOP1 PERFORMANCE)

| Backbone | Accurate Rate (%) under Different Search Size (*10000) | | | |
|---|---|---|---|---|
| | 1 | 3 | 5 | 10 |
| R100-mask-v2 | 89.60 | 85.33 | 83.39 | 80.22 |
| R100-mask-v3 | 89.13 | 84.69 | 82.51 | 79.09 |
| R50-mask-v3 | 84.59 | 78.91 | 76.10 | 71.91 |
| R34-mask | 80.04 | 73.53 | 70.31 | 65.82 |
| Vit-Tiny-mask | 91.42 | 88.64 | 86.55 | 83.69 |
| Vit-Small-mask | 91.04 | 89.01 | 85.96 | 82.55 |

---

[2] TABLE IV has been revised.

TABLE VI. FACE SEARCH RESULT WITH MASK UNDER DIFFERENT SEARCH SIZE (TOP5 PERFORMANCE)

| Backbone | Accurate Rate (%) under Different Search Size (*10000) | | | |
|---|---|---|---|---|
| | 1 | 3 | 5 | 10 |
| R100-mask-v2 | 96.55 | 94.06 | 92.54 | 90.30 |
| R100-mask-v3 | 96.28 | 93.68 | 92.16 | 89.79 |
| R50-mask-v3 | 93.97 | 90.15 | 88.11 | 84.86 |
| R34-mask | 91.14 | 86.38 | 83.83 | 80.04 |
| Vit-Tiny-mask | 97.33 | 95.65 | 93.54 | 90.96 |
| Vit-Small-mask | 97.11 | 95.79 | 92.98 | 91.83 |

In the stage of masked face search, we no longer tested the general models, but only focused on the models designed for masked faces.

We discovered that Vit-Tiny and Vit-Small maintained excellent model accuracy, even in the context of masked face search. Additionally, the large-scale R100 model also demonstrated relatively good search accuracy. However, it still couldn't match the outstanding performance of Vit-Tiny and Vit-Small. This disparity was particularly evident during the masked face tests.

## VI. CONCLUSION

In this work, we explored the impact of different backbones on face recognition, especially in the comparison between masked and unmasked face recognition. In the tests of this masked face recognition model, multiple models were subjected to face comparison and search tests under standard data (100,000 pairs without masks) and masked data (100,000 pairs), and were also compared with state-of-the-art (SOTA) models in the industry.

In the standard data tests, the r100 series of models stood out. When performing face comparison, the accuracy rate exceeded 98% at a one-in-ten-thousand false acceptance rate. In the search tests, both the top1 and top5 metrics were relatively high. Although the accuracy rate decreased as the search database expanded, it still remained at a relatively high level. The r50 series of models performed secondarily, while the r34_mask_v1 model was relatively weaker. In the tests with masked data, the r100_mask_v2 model achieved an accuracy rate of 90.07% in face comparison, leading other models. Among the r50 series of models, the r50_mask_v3 model performed relatively well, yet its overall recognition accuracy was lower than that of the corresponding r100 series models. On the other hand, the Vit-Small and Vit-Tiny models demonstrated good performance, especially in masked face recognition and retrieval, with significant improvements in effectiveness.

However, it must be acknowledged that this work was solely based on the test data of Webface42M, and its effectiveness still needs to be proven through extensive practice. In subsequent work, we will conduct numerous tests and verifications using actual civil aviation face clearance data to ensure the safe and smooth implementation of smart civil aviation and intelligent travel for civil aviation passengers.

ACKNOWLEDGMENT

The authors would like to express their gratitude for the funding from the Open Project of the Key Laboratory of Intelligent Travel for Civil Aviation Passengers of the Civil Aviation Administration of China.

REFERENCES

[1] Lochner, Sabrina A. "Saving face: regulating law enforcement's use of mobile facial recognition technology & iris scans." Ariz. L. Rev. 55 (2013): 201.

[2] Jain, Anil K., Karthik Nandakumar, and Arun Ross. "50 years of biometric research: Accomplishments, challenges, and opportunities." Pattern recognition letters 79 (2016): 80-105.

[3] Taigman, Yaniv, et al. "Deepface: Closing the gap to human-level performance in face verification." Proceedings of the IEEE conference on computer vision and pattern recognition. 2014.

[4] Introna, Lucas, and David Wood. "Picturing algorithmic surveillance: The politics of facial recognition systems." Surveillance & Society 2.2/3 (2004): 177-198.

[5] International Air Transport Association. One ID. IATA, https://www.iata.org/en/programs/passenger/one-id/. Accessed 2 May 2025.

[6] Loey, M., Manogaran, G., Taha, M. H., & Khalifa, N. E. (2021). Fighting against COVID-19: A novel deep learning model based on YOLO-v2 with ResNet-50 for medical face mask detection. Sustainable Cities and Society, 65, 102600.

[7] Kumar, A., Kalia, A., Sharma, A., & Kaushal, M. (2021). A hybrid tiny YOLO v4-SPP module based improved face mask detection vision system. Journal of Ambient Intelligence and Humanized Computing. Advance online publication. https://doi.org/10.1007/s12652-021-03465-2

[8] Mandal, B., Okeukwu, A., & Theis, Y. (2021). Masked face recognition using ResNet-50. arXiv preprint arXiv:2104.08997.

[9] Li, Y., Guo, K., Lu, Y., & Liu, L. (2021). Cropping and attention based approach for masked face recognition. Applied Intelligence, 51(5), 3012–3025.

[10] Boutros, F., Damer, N., Kirchbuchner, F., & Kuijper, A. (2021). Unmasking face embeddings by self-restrained triplet loss for accurate masked face recognition. arXiv preprint arXiv:2103.01716.

[11] Deng, H., Feng, Z., Qian, G., Lv, X., Li, H., & Li, G. (2021). MFCosface: A masked-face recognition algorithm based on large margin cosine loss. Applied Sciences, 11(16), 7310.

[12] Li, C., Ge, S., Zhang, D., & Li, J. (2020). Look through masks: Towards masked face recognition with de-occlusion distillation. In Proceedings of the 28th ACM International Conference on Multimedia (pp. 3016–3024).

[13] Din, N. U., Javed, K., Bae, S., & Yi, J. (2020). A novel GAN-based network for unmasking of masked face. IEEE Access, 8, 44276–44287.

[14] Shahzad, H. M., Bhatti, S. M., Jaffar, A., & Akram, S. (2024). Enhancing masked facial expression recognition with multimodal deep learning. Multimedia Tools and Applications, 83(30), 73911–73921.

[15] Huang, Y. C., Rahardjo, D. A. B., Shiue, R. H., & Chen, H. H. (2024). Masked face recognition using domain adaptation. Pattern Recognition, 153, 110574.

[16] Liao, J., Guha, T., & Sanchez, V. (2025). Self-supervised random mask attention GAN in tackling pose-invariant face recognition. Pattern Recognition, 159, 111112.

[17] LeCun, Yann, et al. "Gradient-based learning applied to document recognition." Proceedings of the IEEE 86.11 (1998): 2278-2324.

[18] Krizhevsky, Alex, Ilya Sutskever, and Geoffrey E. Hinton. "ImageNet classification with deep convolutional neural networks." Communications of the ACM 60.6 (2017): 84-90.

[19] Simonyan, Karen, and Andrew Zisserman. "Very deep convolutional networks for large-scale image recognition." arXiv preprint arXiv:1409.1556 (2014).

[20] Ioffe, Sergey, and Christian Szegedy. "Batch normalization: Accelerating deep network training by reducing internal covariate shift." International conference on machine learning. pmlr, 2015.

[21] Szegedy, Christian, et al. "Rethinking the inception architecture for computer vision." Proceedings of the IEEE conference on computer vision and pattern recognition. 2016.

[22] Dosovitskiy, Alexey, et al. "An image is worth 16x16 words: Transformers for image recognition at scale." arXiv preprint arXiv:2010.11929 (2020).

[23] Wang, Wenhai, et al. "Pyramid vision transformer: A versatile backbone for dense prediction without convolutions." Proceedings of the IEEE/CVF international conference on computer vision. 2021.

[24] Srinivas, Aravind, et al. "Bottleneck transformers for visual recognition." Proceedings of the IEEE/CVF conference on computer vision and pattern recognition. 2021.

[25] Dai, Zihang, et al. "Coatnet: Marrying convolution and attention for all data sizes." Advances in neural information processing systems 34 (2021): 3965-3977.

[26] Zhang, B., et al. "Masked Face Recognition under Different Backbones. " In 2025 IEEE 7th International Conference on Civil Aviation Safety and Information Technology (ICCASIT) (pp. 405-411).